\documentclass[10pt]{llncs}
\usepackage{makeidx}  
\usepackage{graphicx}
\usepackage{multirow}
\usepackage{amsmath}
\usepackage{mathtools}
\DeclarePairedDelimiter\floor{\lfloor}{\rfloor}
\usepackage[]{algorithm2e}
\DeclareGraphicsExtensions{.pdf,.jpeg,.png}
\usepackage{multirow}
\usepackage{array}
\usepackage{tabulary}
\begin{document}

\bibliographystyle{plain}

\frontmatter
\pagestyle{headings}
\mainmatter

\title{Mining Supervisor Evaluation and Peer Feedback in Performance Appraisals}
\titlerunning{Mining Performance Appraisals}

\author{Girish Keshav Palshikar, Sachin Pawar, Saheb Chourasia, Nitin Ramrakhiyani}
\institute{TCS Research, Tata Consultancy Services Limited, \\
54B Hadapsar Industrial Estate, Pune 411013, India.\\
\email{\{gk.palshikar, sachin7.p, saheb.c, nitin.ramrakhiyani\}@tcs.com}}

\maketitle

\begin{abstract}
Performance appraisal (PA) is an important HR process to periodically measure and evaluate every employee's performance vis-a-vis the goals established by the organization. A PA process involves purposeful multi-step multi-modal communication between employees, their supervisors and their peers, such as self-appraisal, supervisor assessment and peer feedback. Analysis of the structured data and text produced in PA is crucial for measuring the quality of appraisals and tracking actual improvements. In this paper, we apply text mining techniques to produce insights from PA text. First, we perform sentence classification to identify strengths, weaknesses and suggestions of improvements found in the supervisor assessments and then use clustering to discover broad categories among them. Next we use multi-class multi-label classification techniques to match supervisor assessments to predefined broad perspectives on performance. Finally, we propose a short-text summarization technique to produce a summary of peer feedback comments for a given employee and compare it with manual summaries. All techniques are illustrated using a real-life dataset of supervisor assessment and peer feedback text produced during the PA of 4528 employees in a large multi-national IT company.
\end{abstract}

\section{Introduction}
\label{s1}

Performance appraisal (PA) is an important HR process, particularly for modern organizations that crucially depend on the skills and expertise of their workforce. The PA process enables an organization to periodically measure and evaluate every employee's performance. It also provides a mechanism to link the goals established by the organization to its each employee's day-to-day activities and performance. Design and analysis of PA processes is a lively area of research within the HR community~\cite{MC95},~\cite{Visw01},~\cite{LW04},~\cite{SBP07}. 

The PA process in any modern organization is nowadays implemented and tracked through an IT system (the {\em PA system}) that records the interactions that happen in various steps. Availability of this data in a computer-readable database opens up opportunities to analyze it using automated statistical, data-mining and text-mining techniques, to generate novel and actionable insights / patterns and to help in improving the quality and effectiveness of the PA process~\cite{PDB09}, ~\cite{RPPA16},~\cite{APPB16}. Automated analysis of large-scale PA data is now facilitated by technological and algorithmic advances, and is becoming essential for large organizations containing thousands of geographically distributed employees handling a wide variety of roles and tasks. 

A typical PA process involves purposeful multi-step multi-modal communication between employees, their supervisors and their peers. In most PA processes, the communication includes the following steps: (i) in {\em self-appraisal}, an employee records his/her achievements, activities, tasks handled etc.; (ii) in {\em supervisor assessment}, the supervisor provides the criticism, evaluation and suggestions for improvement of performance etc.; and (iii) in {\em peer feedback} (aka {\em $360^\circ$ view}), the peers of the employee provide their feedback. There are several business questions that managers are interested in. Examples: 
\begin{enumerate}
\item For my workforce, what are the broad categories of strengths, weaknesses and suggestions of improvements found in the supervisor assessments? 
\item For my workforce, how many supervisor comments are present for each of a given fixed set of perspectives (which we call {\em attributes}), such as {\small {\sf FUNCTIONAL\_EXCELLENCE}}, {\small {\sf CUSTOMER\_FOCUS}}, {\small {\sf BUILDING\_EFFECTIVE\_TEAMS}} etc.?
\item What is the summary of the peer feedback for a given employee? 
\end{enumerate}
In this paper, we develop text mining techniques that can automatically produce answers to these questions. Since the intended users are HR executives, ideally, the techniques should work with minimum training data and experimentation with parameter setting. These techniques have been implemented and are being used in a PA system in a large multi-national IT company. 

The rest of the paper is organized as follows. Section~\ref{s2} summarizes related work. Section~\ref{s3} summarizes the PA dataset used in this paper. Section~\ref{s4} applies sentence classification algorithms to automatically discover three important classes of sentences in the PA corpus viz., sentences that discuss strengths, weaknesses of employees and contain suggestions for improving her performance. Section~\ref{s5} considers the problem of mapping the actual targets mentioned in strengths, weaknesses and suggestions to a fixed set of attributes. In Section~\ref{s6}, we discuss how the feedback from peers for a particular employee can be summarized. In Section~\ref{s7} we draw conclusions and identify some further work. 

\section{Related Work}\label{s2}

We first review some work related to sentence classification. Semantically classifying sentences (based on the sentence's purpose) is a much harder task, and is gaining increasing attention from linguists and NLP researchers. McKnight and Srinivasan~\cite{MS03} and Yamamoto and Takagi~\cite{YT05} used SVM to classify sentences in biomedical abstracts into  classes such as {\small {\sf INTRODUCTION, BACKGROUND, PURPOSE, METHOD, RESULT, CONCLUSION}}. 
Cohen et al.~\cite{CCM04} applied SVM and other techniques to learn classifiers for sentences in emails into classes, which are speech acts defined by a verb-noun pair, with verbs such as {\small {\tt request, propose, amend, commit, deliver}} and nouns such as {\small {\tt meeting, document, committee}}; see also ~\cite{CC06}. Khoo et al.~\cite{KMA06} uses various classifiers to classify sentences in emails into classes such as {\small {\sf APOLOGY, INSTRUCTION, QUESTION, REQUEST, SALUTATION, STATEMENT, SUGGESTION, THANKING}} etc. 
Qadir and Riloff~\cite{QR11} proposes several filters and classifiers to classify sentences on message boards (community QA systems) into 4 speech acts: {\small {\sf COMMISSIVE}} (speaker commits to a future action),  {\small {\sf DIRECTIVE}} (speaker expects listener to take some action), {\small {\sf EXPRESSIVE}} (speaker expresses his or her psychological state to the listener), {\small {\sf REPRESENTATIVE}} (represents the speaker's belief of something). Hachey and Grover~\cite{HG05} used SVM and maximum entropy classifiers to classify sentences in legal documents into classes such as {\small {\sf FACT, PROCEEDINGS, BACKGROUND, FRAMING, DISPOSAL}}; see also~\cite{RPP13}. Deshpande et al.~\cite{DPA10} proposes unsupervised linguistic patterns to classify sentences into classes {\small {\sf SUGGESTION, COMPLAINT}}. 

There is much work on a closely related problem viz., classifying sentences in dialogues through dialogue-specific categories called {\em dialogue acts}~\cite{SRCS00}, which we will not review here. Just as one example, Cotterill~\cite{Cott11} classifies questions in emails into the dialogue acts of {\small {\sf YES\_NO\_QUESTION, WH\_QUESTION, ACTION\_REQUEST, RHETORICAL, MULTIPLE\_CHOICE}} etc.   

We could not find much work related to mining of performance appraisals data. Pawar et al.~\cite{PRPH15} uses kernel-based classification to classify sentences in both performance appraisal text and product reviews into classes {\small {\sf SUGGESTION, APPRECIATION, COMPLAINT}}. Apte et al.~\cite{APPB16} provides two algorithms for matching the descriptions of goals or tasks assigned to employees to a standard template of model goals. One algorithm is based on the co-training framework and uses goal descriptions and self-appraisal comments as two separate perspectives. The second approach uses semantic similarity under a weak supervision framework. Ramrakhiyani et al.~\cite{RPPA16} proposes label propagation algorithms to discover aspects in supervisor assessments in performance appraisals, where an aspect is modelled as a verb-noun pair (e.g. {\small {\tt conduct training, improve coding}}).

\section{Dataset}\label{s3}

In this paper, we used the supervisor assessment and peer feedback text produced during the performance appraisal of 4528 employees in a large multi-national IT company. The corpus of supervisor assessment has 26972 sentences. The summary statistics about the number of words in a sentence is: min:4 max:217 average:15.5 STDEV:9.2 Q1:9 Q2:14 Q3:19. 

\section{Sentence Classification}\label{s4} 

The PA corpus contains several classes of sentences that are of interest. In this paper, we focus on three important classes of sentences viz., sentences that discuss strengths (class {\small {\sf STRENGTH}}), weaknesses of employees (class {\small {\sf WEAKNESS}}) and suggestions for improving her performance (class {\small {\sf SUGGESTION}}). The strengths or weaknesses are mostly about the performance in work carried out, but sometimes they can be about the working style or other personal qualities. The classes {\small {\sf WEAKNESS}} and {\small {\sf SUGGESTION}} are somewhat overlapping; e.g., a suggestion may address a perceived weakness. Following are two example sentences in each class.\\*
\\*

\noindent{\small {\sf STRENGTH}}:
\begin{itemize}
\item {\small {\tt Excellent technology leadership and delivery capabilities along with ability to groom technology champions within the team.}}
\item {\small {\tt He can drive team to achieve results and can take pressure.}}
\end{itemize}

\noindent{\small {\sf WEAKNESS}}:
\begin{itemize}
\item {\small {\tt Sometimes exhibits the quality that he knows more than the others in the room which puts off others.}}
\item {\small {\tt Tends to stretch himself and team a bit too hard.}}
\end{itemize}

\noindent{\small {\sf SUGGESTION}}:
\begin{itemize}
\item {\small {\tt X has to attune himself to the vision of the business unit and its goals a little more than what is being currently exhibited.}}
\item {\small {\tt Need to improve on business development skills, articulation of business and solution benefits.}}
\end{itemize}

Several linguistic aspects of these classes of sentences are apparent. The subject is implicit in many sentences. The strengths are often mentioned as either noun phrases (NP) with positive adjectives ({\small {\tt Excellent technology leadership}}) or positive nouns ({\small {\tt engineering strength}}) or through verbs with positive polarity ({\small {\tt dedicated}}) or as verb phrases containing positive adjectives ({\small {\tt delivers innovative solutions}}). Similarly for weaknesses, where negation is more frequently used ({\small {\tt presentations are not his forte}}), or alternatively, the polarities of verbs ({\small {\tt avoid}}) or adjectives ({\small {\tt poor}}) tend to be negative. However, sometimes the form of both the strengths and weaknesses is the same, typically a stand-alone sentiment-neutral NP, making it difficult to distinguish between them; e.g., {\small {\tt adherence to timing}} or {\small {\tt timely closure}}. Suggestions often have an imperative mood and contain secondary verbs such as {\small {\tt need to, should, has to}}. Suggestions are sometimes expressed using comparatives ({\small {\tt better process compliance}}). We built a simple set of patterns for each of the 3 classes on the POS-tagged form of the sentences. We use each set of these patterns as an unsupervised sentence classifier for that class. If a particular sentence matched with patterns for multiple classes, then we have simple tie-breaking rules for picking the final class.  
The pattern for the {\small {\sf STRENGTH}} class looks for the presence of positive words / phrases like {\small {\tt takes ownership}}, {\small {\tt excellent}}, {\small {\tt hard working}}, {\small {\tt commitment}}, etc. Similarly, the pattern for the {\small {\sf WEAKNESS}} class looks for the presence of negative words / phrases like {\small {\tt lacking}}, {\small {\tt diffident}}, {\small {\tt slow learner}}, {\small {\tt less focused}}, etc. The {\small {\sf SUGGESTION}} pattern not only looks for keywords like {\small {\tt should}}, {\small {\tt needs to}} but also for POS based pattern like ``a verb in the base form (VB) in the beginning of a sentence''. 

We randomly selected 2000 sentences from the supervisor assessment corpus and manually tagged them (dataset D1). This labelled dataset contained 705, 103, 822 and 370 sentences having the class labels {\small {\sf STRENGTH}}, {\small {\sf WEAKNESS}}, {\small {\sf SUGGESTION}} or {\small {\sf OTHER}} respectively. 
We trained several multi-class classifiers on this dataset. Table~\ref{t1} shows the results of 5-fold cross-validation experiments on dataset D1. For the first 5 classifiers, we used their implementation from the SciKit Learn library in Python ({\small {\tt scikit-learn.org}}). The features used for these classifiers were simply the sentence words along with their frequencies. For the last 2 classifiers (in Table~\ref{t1}), we used our own implementation. The overall {\em accuracy} for a classifier is defined as $A = \frac{\#correct\_predictions}{\#data\_points}$, where the denominator is 2000 for dataset D1. Note that the pattern-based approach is unsupervised i.e., it did not use any training data. Hence, the results shown for it are for the entire dataset and not based on cross-validation. 

\begin{table}
\caption{Results of 5-fold cross validation for sentence classification on dataset D1.}
\label{t1}
\small
\centering
\begin{tabular}{|l||c|c|c||c|c|c||c|c|c||c|}
\hline
                           & \multicolumn{3}{|c||}{\textbf{\small {\sf STRENGTH}}} & \multicolumn{3}{|c||}{\textbf{\small {\sf WEAKNESS}}} & \multicolumn{3}{|c||}{\textbf{\small {\sf SUGGESTION}}} & \\
\hline
\textbf{Classifier} & \textbf{P} & \textbf{R} & \textbf{F} & \textbf{P} & \textbf{R} & \textbf{F} & \textbf{P} & \textbf{R} & \textbf{F} & \textbf{A}\\
\hline
\hline
Logistic Regression & 0.715 & 0.759 & 0.736 & 0.309 & 0.204 & 0.246 & 0.788 & 0.749 & 0.768 & 0.674\\
Multinomial Naive Bayes & 0.719 & 0.723 & 0.721 & 0.246 & 0.155 & 0.190 & 0.672 & 0.790 & 0.723 & 0.646\\
Random Forest & 0.681 & 0.688 & 0.685 & 0.286 & 0.039 & 0.068 & 0.730 & 0.734 & 0.732 & 0.638\\
AdaBoost & 0.522 & 0.888 & 0.657 & 0.265 & 0.087 & 0.131 & 0.825 & 0.618 & 0.707 & 0.604\\
Linear SVM & 0.718 & 0.698 & 0.708 & 0.357 & 0.194 & 0.252 & 0.744 & 0.759 & 0.751 & 0.651\\
SVM with ADWSK~\cite{PRPH15} & 0.789 & 0.847 & \textbf{0.817} & 0.491 & 0.262 & 0.342 & 0.844 & 0.871 & \textbf{0.857} & \textbf{0.771}\\
Pattern-based & 0.825 & 0.687 & 0.749 & 0.976 & 0.494 & \textbf{0.656} & 0.835 & 0.828 & 0.832 & 0.698\\ 
\hline
\end{tabular}
\end{table}
\normalsize

\subsection{Comparison with Sentiment Analyzer}
We also explored whether a sentiment analyzer can be used as a baseline for identifying the class labels {\small {\sf STRENGTH}} and {\small {\sf WEAKNESS}}. We used an implementation of sentiment analyzer from TextBlob\footnote{\url{https://textblob.readthedocs.io/en/dev/}} to get a polarity score for each sentence. Table~\ref{tab:sentiment} shows the distribution of positive, negative and neutral sentiments across the 3 class labels {\small {\sf STRENGTH}}, {\small {\sf WEAKNESS}} and {\small {\sf SUGGESTION}}. It can be observed that distribution of positive and negative sentiments is almost similar in {\small {\sf STRENGTH}} as well as {\small {\sf SUGGESTION}} sentences, hence we can conclude that the information about sentiments is not much useful for our classification problem.

\begin{table}
\caption{Results of TextBlob sentiment analyzer on the dataset D1}
\label{tab:sentiment}
\centering
\begin{tabular}{|c|c|c|c|}
\hline
\textbf{Sentence Class} & \textbf{Positive} & \textbf{Negative} & \textbf{Neutral}\\
\hline
{\small {\sf STRENGTH}} & 544 & 44 & 117 \\
{\small {\sf WEAKNESS}} & 44 & 24 & 35 \\
{\small {\sf SUGGESTION}} & 430 & 52 & 340 \\
\hline
\end{tabular}
\end{table}

\subsection{Discovering Clusters within Sentence Classes}
\begin{table}[tbp]
\caption{5 representative clusters in strengths.}
\label{t2s}
\small
\centering
\begin{tabular}{|l|c|}
\hline
\textbf{Strength cluster} & \textbf{Count}\\
\hline
{\tt motivation expertise knowledge talent skill} & 1851 \\
{\tt coaching team coach} & 1787 \\
{\tt professional career job work working training practice} & 1531 \\
{\tt opportunity focus attention success future potential impact result change} & 1431 \\
{\tt sales retail company business industry marketing product} & 1251 \\
\hline
\end{tabular}
\end{table}
\normalsize


\begin{table}[tbp]
\caption{5 representative clusters in weaknesses and suggestions.}
\label{t2w}
\small
\centering
\begin{tabular}{|l|c|}
\hline
\textbf{Weakness cluster} & \textbf{Count}\\
\hline
{\tt motivation expertise knowledge talent skill} & 1308\\
{\tt market sales retail corporate marketing commercial industry business} & 1165\\
{\tt awareness emphasis focus} & 1165\\
{\tt coaching team coach} & 1149\\
{\tt job work working task planning} & 1074\\
\hline
\end{tabular}
\end{table}
\normalsize

After identifying sentences in each class, we can now answer question (1) in Section~\ref{s1}. From 12742 sentences predicted to have label {\small {\sf STRENGTH}}, we extract nouns that indicate the actual strength, and cluster them using a simple clustering algorithm which uses the cosine similarity between word embeddings\footnote{We used 100 dimensional word vectors trained on Wikipedia 2014 and Gigaword 5 corpus, available at: \url{https://nlp.stanford.edu/projects/glove/}} of these nouns. We repeat this for the 9160 sentences with predicted label {\small {\sf WEAKNESS}} or {\small {\sf SUGGESTION}} as a single class. Tables~\ref{t2s} and~\ref{t2w} show a few representative clusters in strengths and in weaknesses, respectively. 
We also explored clustering 12742 {\small {\sf STRENGTH}} sentences directly using CLUTO~\cite{karypis2002cluto} and Carrot2 Lingo~\cite{DBLP:conf/iis/OsinskiSW04} clustering algorithms. Carrot2 Lingo\footnote{We used the default parameter settings for Carrot2 Lingo algorithm as mentioned at: \url{http://download.carrot2.org/head/manual/index.html}} discovered 167 clusters and also assigned labels to these clusters. We then generated 167 clusters using CLUTO as well. CLUTO does not generate cluster labels automatically, hence we used 5 most frequent words within the cluster as its labels. Table~\ref{clutocarrot} shows the largest 5 clusters by both the algorithms. It was observed that the clusters created by CLUTO were more meaningful and informative as compared to those by Carrot2 Lingo. Also, it was observed that there is some correspondence between noun clusters and sentence clusters. E.g. the nouns cluster \texttt{motivation expertise knowledge talent skill} (Table~\ref{t2s}) corresponds to the CLUTO sentence cluster \texttt{skill customer management knowledge team} (Table~\ref{clutocarrot}). But overall, users found the nouns clusters to be more meaningful than the sentence clusters.

\begin{table}
\caption{Largest 5 sentence clusters within 12742 {\small {\sf STRENGTH}} sentences}
\label{clutocarrot}
\begin{tabular}{|c|l|c|}
\hline
\textbf{Algorithm} & \textbf{Cluster} & \textbf{\#Sentences} \\
\hline
\multirow{5}{*}{CLUTO} & {\tt performance performer perform years team} & 510 \\
\cline{2-3}
 & {\tt skill customer management knowledge team} & 325 \\ 
\cline{2-3}
 & {\tt role delivery work place show} & 289 \\
\cline{2-3}
 & {\tt delivery manage management manager customer} & 259 \\
\cline{2-3}
 & {\tt knowledge customer business experience work} & 250 \\
\hline
\multirow{5}{*}{Carrot2} & {\tt manager manage} & 1824 \\
\cline{2-3}
 & {\tt team team} & 1756 \\ 
\cline{2-3}
 & {\tt delivery management} & 451 \\
\cline{2-3}
 & {\tt manage team} & 376 \\
\cline{2-3}
 & {\tt customer management} & 321 \\
\hline
\end{tabular}
\end{table}

\section{PA along Attributes}\label{s5}

In many organizations, PA is done from a predefined set of perspectives, which we call {\em attributes}. Each attribute covers one specific aspect of the work done by the employees. This has the advantage that we can easily compare the performance of any two employees (or groups of employees) along any given attribute. We can correlate various performance attributes and find dependencies among them. We can also cluster employees in the workforce using their supervisor ratings for each attribute to discover interesting insights into the workforce. The HR managers in the organization considered in this paper have defined 15 attributes (Table~\ref{t3}). Each attribute is essentially a work item or work category described at an abstract level. For example, {\small {\sf FUNCTIONAL\_EXCELLENCE}} covers any tasks, goals or activities related to the software engineering life-cycle (e.g., requirements analysis, design, coding, testing etc.) as well as technologies such as databases, web services and GUI. 

\begin{table}[tbp]
\caption{Strengths, Weaknesses and Suggestions along Performance Attributes}
\label{t3}
\small
\centering
\begin{tabular}{|l|c|c|c|}
\hline
\textbf{Performance Attributes} & \textbf{\#Strengths} & \textbf{\#Weaknesses} & \textbf{\#Suggestions}\\
\hline
{\small {\sf FUNCTIONAL\_EXCELLENCE}} & 321 & 26 & 284\\
{\small {\sf BUILDING\_EFFECTIVE\_TEAMS}} & 80 & 6 & 89\\
{\small {\sf INTERPERSONAL\_EFFECTIVENESS}} & 151 & 16 & 97\\
{\small {\sf CUSTOMER\_FOCUS}} & 100 & 5 & 76\\
{\small {\sf INNOVATION\_MANAGEMENT}} & 22 & 4 & 53\\
{\small {\sf EFFECTIVE\_COMMUNICATION}} & 53 & 17 & 124\\
{\small {\sf BUSINESS\_ACUMEN}} & 39 & 10 & 103\\
{\small {\sf TAKING\_OWNERSHIP}} & 47 & 3 & 81\\
{\small {\sf PEOPLE\_DEVELOPMENT}} & 31 & 8 & 57\\
{\small {\sf DRIVE\_FOR\_RESULTS}} & 37 & 4 & 30\\
{\small {\sf STRATEGIC\_CAPABILITY}} & 8 & 4 & 51\\
{\small {\sf WITHSTANDING\_PRESSURE}} & 16 & 6 & 16\\
{\small {\sf DEALING\_WITH\_AMBIGUITIES}} & 4 & 8 & 12\\
{\small {\sf MANAGING\_VISION\_AND\_PURPOSE}} & 3 & 0 & 9\\
{\small {\sf TIMELY\_DECISION\_MAKING}} & 6 & 2 & 10\\
\hline
\end{tabular}
\end{table}
\normalsize

In the example in Section~\ref{s4}, the first sentence (which has class {\small {\sf STRENGTH}}) can be mapped to two attributes: {\small {\sf FUNCTIONAL\_EXCELLENCE}} and {\small {\sf BUILDING\_EFFECTIVE\_TEAMS}}. Similarly, the third sentence (which has class {\small {\sf WEAKNESS}}) can be mapped to the attribute  {\small {\sf INTERPERSONAL\_EFFECTIVENESS}} and so forth. Thus, in order to answer the second question in Section~\ref{s1}, we need to map each sentence in each of the 3 classes to zero, one, two or more attributes, which is a multi-class multi-label classification problem. 

We manually tagged the same 2000 sentences in Dataset D1 with attributes, where each sentence may get 0, 1, 2, etc. up to 15 class labels (this is dataset D2). This labelled dataset contained 749, 206, 289, 207, 91, 223, 191, 144, 103, 80, 82, 42, 29, 15, 24 sentences having the class labels listed in Table~\ref{t3} in the same order. The number of sentences having 0, 1, 2, or more than 2 attributes are: 321, 1070, 470 and 139 respectively. We trained several multi-class multi-label classifiers on this dataset. Table~\ref{t4} shows the results of 5-fold cross-validation experiments on dataset D2.

Precision, Recall and F-measure for this multi-label classification are computed using a strategy similar to the one described in~\cite{godbole2004}. Let $P_i$ be the set of predicted labels and $A_i$ be the set of actual labels for the $i^{th}$ instance. Precision and recall for this instance are computed as follows:
\begin{equation*}
Precision_i = \frac{|P_i \cap A_i|}{|P_i|},\hspace{2mm}
Recall_i = \frac{|P_i \cap A_i|}{|A_i|}
\end{equation*}
It can be observed that $Precision_i$ would be undefined if $P_i$ is empty and similarly $Recall_i$ would be undefined when $A_i$ is empty. Hence, overall precision and recall are computed by averaging over all the instances except where they are undefined. Instance-level F-measure can not be computed for instances where either precision or recall are undefined. Therefore, overall F-measure is computed using the overall precision and recall.

\begin{table}[tbp]
\caption{Results of 5-fold cross validation for multi-class multi-label classification on dataset D2.}
\label{t4}
\small
\centering
\begin{tabular}{|l|c|c|c|}
\hline
\textbf{Classifier} & \textbf{Precision} \textbf{P} & \textbf{Recall} \textbf{R} & \textbf{F}\\
\hline
Logistic Regression & 0.715 & 0.711 & \textbf{0.713}\\
Multinomial Naive Bayes & 0.664 & 0.588 & 0.624\\
Random Forest & 0.837 & 0.441 & 0.578\\
AdaBoost & 0.794 & 0.595 & 0.680\\
Linear SVM & 0.722 & 0.672 & 0.696\\
Pattern-based & 0.750 & 0.679 & \textbf{0.713}\\
\hline
\end{tabular}
\end{table}
\normalsize

\section{Summarization of Peer Feedback using ILP}\label{s6}

The PA system includes a set of peer feedback comments for each employee. To answer the third question in Section~\ref{s1}, we need to create a summary of all the peer feedback comments about a given employee. As an example, following are the feedback comments from 5 peers of an employee.

\begin{enumerate}
\item {\small {\tt vast knowledge on different technologies}}
\item {\small {\tt His experience and wast knowledge mixed with his positive attitude, willingness to teach and listen and his humble nature.}}
\item {\small {\tt Approachable, Knowlegeable and is of helping nature.}}
\item {\small {\tt Dedication, Technical expertise and always supportive}}
\item {\small {\tt Effective communication and team player}}
\end{enumerate}

The individual sentences in the comments written by each peer are first identified and then POS tags are assigned to each sentence. We hypothesize that a good summary of these multiple comments can be constructed by identifying a set of {\em important} text fragments or phrases. Initially, a set of candidate phrases is extracted from these comments and a subset of these candidate phrases is chosen as the final summary, using Integer Linear Programming (ILP). The details of the ILP formulation are shown in Table~\ref{ilp}. As an example, following is the summary generated for the above 5 peer comments.

\noindent{\small {\tt humble nature, effective communication, technical expertise, always supportive, vast knowledge}}

\vspace{2mm}
\noindent Following rules are used to identify candidate phrases:
\begin{itemize}
\item An adjective followed by \texttt{in} which is followed by a noun phrase (e.g. \texttt{good in customer relationship})
\item A verb followed by a noun phrase (e.g. \texttt{maintains work life balance})
\item A verb followed by a preposition which is followed by a noun phrase (e.g. \texttt{engage in discussion})
\item Only a noun phrase (e.g. \texttt{excellent listener})
\item Only an adjective (e.g. \texttt{supportive})
\end{itemize}

\noindent Various parameters are used to evaluate a candidate phrase for its {\em importance}. A candidate phrase is more important:
\begin{itemize}
\item if it contains an adjective or a verb or its headword is a noun having WordNet lexical category {\em noun.attribute} (e.g. nouns such as \texttt{dedication}, \texttt{sincerity})
\item if it contains more number of words
\item if it is included in comments of multiple peers
\item if it represents any of the performance attributes such as {\em Innovation}, {\em Customer}, {\em Strategy} etc.
\end{itemize}
A complete list of parameters is described in detail in Table~\ref{ilp}.

There is a trivial constraint $C_0$ which makes sure that only $K$ out of $N$ candidate phrases are chosen. A suitable value of $K$ is used for each employee depending on number of candidate phrases identified across all peers (see Algorithm~\ref{alg:K}). Another set of constraints ($C_1$ to $C_{10}$) make sure that at least one phrase is selected for each of the leadership attributes. The constraint $C_{11}$ makes sure that multiple phrases sharing the same headword are not chosen at a time. Also, single word candidate phrases are chosen only if they are adjectives or nouns with lexical category {\em noun.attribute}. This is imposed by the constraint $C_{12}$. It is important to note that all the constraints except $C_0$ are soft constraints, i.e. there may be feasible solutions which do not satisfy some of these constraints. But each constraint which is not satisfied, results in a penalty through the use of slack variables. These constraints are described in detail in Table~\ref{ilp}.

The objective function maximizes the total {\em importance} score of the selected candidate phrases. At the same time, it also minimizes the sum of all slack variables so that the minimum number of constraints are broken.

\begin{algorithm}\footnotesize
 \KwData{$N$: No. of candidate phrases}
 \KwResult{$K$: No. of phrases to select as part of summary}
 
  \uIf{$N\leq 10$}{
   $K\gets \floor*{N*0.5}$\;
   }
   \uElseIf{$N\leq 20$}{
   $K\gets \floor*{\hspace{1mm}getNoOfPhrasesToSelect(10)+ {(N-10)*0.4}}$\;
  }
  \uElseIf{$N\leq 30$}{
  $K\gets \floor*{\hspace{1mm}getNoOfPhrasesToSelect(20)+ {(N-20)*0.3}}$\;
  }
  \uElseIf{$N\leq 50$}{
  $K\gets \floor*{\hspace{1mm}getNoOfPhrasesToSelect(30)+ {(N-30)*0.2}}$\;
  }
  \Else{
  $K\gets \floor*{\hspace{1mm}getNoOfPhrasesToSelect(50)+ {(N-50)*0.1}}$\;
  }
  
  \uIf{$K < 4$ and $N\geq 4$}{
  $K\gets 4$
  }
  \uElseIf{$K < 4$}{
  $K\gets N$
  }
  \ElseIf{$K > 20$}{
  $K\gets 20$
  }

 \caption{$getNoOfPhrasesToSelect$ (For determining number of phrases to select to include in summary)}
 \label{alg:K}
\end{algorithm}

%
%
%

\begin{table}
\caption{Integer Linear Program (ILP) formulation}
\label{ilp}
\begin{tabular}{|p{0.95\columnwidth}|}
\hline
\textbf{Parameters}:
\begin{itemize}
\item $N$: No. of phrases
\item $K$: No. of phrases to be chosen for inclusion in the final summary
\item $Freq$: Array of size $N$, $Freq_i=$ no. of distinct peers mentioning the $i^{th}$ phrase
\item $Adj$: Array of size $N$, $Adj_i=1$ if the $i^{th}$ phrase contains any adjective
\item $Verb$: Array of size $N$, $Verb_i=1$ if the $i^{th}$ phrase contains any verb
\item $NumWords$: Array of size $N$, $NumWords_i=1$ no. of words in the $i^{th}$ phrase
\item $NounCat$: Array of size $N$, $NounCat_i=1$ if lexical category (WordNet) of headword of the $i^{th}$ phrase is {\em noun.attribute}
\item $InvalidSingleNoun$: Array of size $N$, $InvalidSingleNoun_i=1$ if the $i^{th}$ phrase is single word phrase which is neither an adjective nor a noun having lexical category (WordNet) {\em noun.attribute}
\item $Leadership, Team, Innovation, Communication, Knowledge, Delivery,$ $Ownership, Customer, Strategy, Personal$: Indicator arrays of size $N$ each, representing whether any phrase corresponds to a particular performance attribute, e.g. $Customer_i=1$ indicates that $i^{th}$ phrase is of type $Customer$
\item $S$: Matrix of dimensions $N\times N$, where $S_{ij}=1$ if headwords of $i^{th}$ and $j^{th}$ phrase are same
\end{itemize}\\
\hline
\textbf{Variables}:
\begin{itemize}
\item $X$: Array of $N$ \textbf{binary} variables, where $X_{i}=1$ only when $i^{th}$ phrase is chosen to be the part of final summary
\item $S_1, S_2, \cdots S_{12}$: \textbf{Integer} slack variables
\end{itemize}\\
\hline
\textbf{Objective}:\\
Maximize $\sum_{i=1}^{N} \left(\left( NounCat_i + Adj_i + Verb_i + 1 \right) \cdot Freq_i \cdot NumWords_i \cdot X_i\right)$\\
\hspace{15mm}$-10000\cdot \sum_{j=1}^{12}S_j$\\
\hline
\textbf{Constraints}:\\
\hspace{2mm}$C_0$: $\sum_{i=1}^{N}X_i = K$ (Exactly $K$ phrases should be chosen)\\
\vspace{1mm}
\hspace{2mm}$C_1$: $\sum_{i=1}^{N} (Leadership_i\cdot X_i) + S_1 \geq 1$\\
\hspace{2mm}$C_2$: $\sum_{i=1}^{N} (Team_i\cdot X_i) + S_2 \geq 1$\\
\hspace{2mm}$C_3$: $\sum_{i=1}^{N} (Knowledge_i\cdot X_i) + S_3 \geq 1$\\
\hspace{2mm}$C_4$: $\sum_{i=1}^{N} (Delivery_i\cdot X_i) + S_4 \geq 1$\\
\hspace{2mm}$C_5$: $\sum_{i=1}^{N} (Ownership_i\cdot X_i) + S_5 \geq 1$\\
\hspace{2mm}$C_6$: $\sum_{i=1}^{N} (Innovation_i\cdot X_i) + S_6 \geq 1$\\
\hspace{2mm}$C_7$: $\sum_{i=1}^{N} (Communication_i\cdot X_i) + S_7 \geq 1$\\
\hspace{2mm}$C_8$: $\sum_{i=1}^{N} (Customer_i\cdot X_i) + S_8 \geq 1$\\
\hspace{2mm}$C_9$: $\sum_{i=1}^{N} (Strategy_i\cdot X_i) + S_9 \geq 1$\\
\hspace{2mm}$C_{10}$: $\sum_{i=1}^{N} (Personal_i\cdot X_i) + S_10 \geq 1$\\
\hspace{4mm}(At least one phrase should be chosen to represent each leadership attribute)\\
\vspace{1mm}
\hspace{2mm}$C_{11}$: $\sum_{i=1}^N\sum_{j=1, s.t. i\neq j}^N (S_{ij}\cdot (X_i + X_j - 1)) + S_{11} <= 0$\\
\hspace{4mm}(No duplicate phrases should be chosen)\\
\vspace{1mm}
\hspace{2mm}$C_{12}$: $\sum_{i=1}^N (InvalidSingleNoun_i \cdot  X_i) - S_{12} <= 0$\\
\hspace{4mm}(Single word noun phrases are not preferred if they are not {\em noun.attribute})\\
\hline
\end{tabular}
\end{table}

\subsection{Evaluation of auto-generated summaries}
We considered a dataset of 100 employees, where for each employee multiple peer comments were recorded. Also, for each employee, a manual summary was generated by an HR personnel. The summaries generated by our ILP-based approach were compared with the corresponding manual summaries using the ROUGE~\cite{lin2004rouge} unigram score. 
For comparing performance of our ILP-based summarization algorithm, we explored a few summarization algorithms provided by the Sumy package\footnote{\url{https://github.com/miso-belica/sumy}}. A common parameter which is required by all these algorithms is number of sentences keep in the final summary. ILP-based summarization requires a similar parameter K, which is automatically decided based on number of total candidate phrases. Assuming a sentence is equivalent to roughly 3 phrases, for Sumy algorithms, we set number of sentences parameter to the ceiling of K/3. Table~\ref{summaryresults} shows average and standard deviation of ROUGE unigram f1 scores for each algorithm, over the 100 summaries. The performance of ILP-based summarization is comparable with the other algorithms, as the two sample t-test does not show statistically significant difference. Also, human evaluators preferred phrase-based summary generated by our approach to the other sentence-based summaries.


\begin{table}
\caption{Comparative performance of various summarization algorithms}
\label{summaryresults}
\centering
\begin{tabular}{|c|c|c|}
\hline
\multirow{2}{*}{Algorithm} & \multicolumn{2}{c|}{ROUGE unigram F1}\\
\cline{2-3}
 & Average & Std. Deviation\\
\hline
LSA & 0.254 & 0.146 \\
TextRank & 0.254 & 0.146 \\
LexRank & 0.258 & 0.148 \\
ILP-based summary & 0.243 & 0.15 \\
\hline
\end{tabular}
\end{table}

\section{Conclusions and Further Work}\label{s7}
In this paper, we presented an analysis of the text generated in Performance Appraisal (PA) process in a large multi-national IT company. We performed sentence classification to identify strengths, weaknesses and suggestions for improvements found in the supervisor assessments and then used clustering to discover broad categories among them. As this is non-topical classification, we found that SVM with ADWS kernel~\cite{PRPH15} produced the best results. We also used multi-class multi-label classification techniques to match supervisor assessments to predefined broad perspectives on performance. Logistic Regression classifier was observed to produce the best results for this topical classification. Finally, we proposed an ILP-based summarization technique to produce a summary of peer feedback comments for a given employee and compared it with manual summaries. 

The PA process also generates much structured data, such as supervisor ratings. It is an interesting problem to compare and combine the insights from discovered from structured data and unstructured text. Also, we are planning to automatically discover any additional performance attributes to the list of 15 attributes currently used by HR.

\bibliographystyle{splncs}
\bibliography{Draft1}

\end{document}